\documentclass[a4paper,final]{article}
\usepackage[dvips]{graphicx}
\usepackage{amsmath} 
\usepackage{amssymb} 
\renewenvironment{abstract}
  {\normalfont
    \list{}{\labelwidth0pt
      \leftmargin0pt \rightmargin\leftmargin
      \listparindent\parindent \itemindent0pt
      \parsep0pt
      
    }
    \item[\hskip\labelsep\bfseries\abstractname\enspace --] \itshape
}{
  \endlist}

\newcommand{\keywordsname}{Keywords}
\newenvironment{keywords}
  {\normalfont
    \list{}{\labelwidth0pt
      \leftmargin0pt \rightmargin\leftmargin
      \listparindent\parindent \itemindent0pt
      \parsep0pt
      }
    \item[\hskip\labelsep\bfseries\keywordsname:]}{\endlist}

\begin{document}

\pagestyle{myheadings}
\markboth{}{}

\title{Target Type Tracking with PCR5 and Dempster's rules:\\
A Comparative Analysis\thanks {This work is partially supported by the Bulgarian
National Science Fund- grants I-1202/02 , 
I-1205/02, MI-1506/05, EC FP6 funded project - BIS21++
(FP6-2004-ACC-SSA-2)}
}

\author{\begin{tabular}{c@{\extracolsep{8em}}c}
\\
Jean Dezert & Albena Tchamova \\
ONERA & IPP, Bulgarian Academy of Sciences\\
29 Av. de la  Division Leclerc & "Acad. G. Bonchev" Str., bl. 25-A\\
92320 Ch\^{a}tillon, France & 1113 Sofia, Bulgaria\\
{\tt Jean.Dezert@onera.fr} & {\tt tchamova@bas.bg}\\
\\
Florentin Smarandache & Pavlina Konstantinova\\
Department of Mathematics & IPP, Bulgarian Academy of Sciences\\
University of New Mexico & "Acad. G. Bonchev" Str., bl. 25-A\\
Gallup, NM 87301, U.S.A. & 1113 Sofia, Bulgaria\\
{\tt smarand@unm.edu} & {\tt pavlina@bas.bg}
\end{tabular}}

\date{}

\maketitle

\begin{abstract}
In this paper we consider and analyze the behavior of two combinational rules for temporal (sequential) attribute data fusion for target type estimation. Our comparative analysis is based on Dempster's fusion rule proposed in Dempster-Shafer Theory (DST) and on the Proportional Conflict Redistribution rule no. 5 (PCR5) recently proposed in Dezert-Smarandache Theory (DSmT).  We show through very simple scenario and Monte-Carlo simulation, how PCR5 allows a very efficient Target Type Tracking and reduces drastically the latency delay for correct Target Type decision with respect to Demspter's rule. For cases presenting some short Target Type switches, Demspter's rule is proved to be unable to detect the switches and thus to track correctly the Target Type changes. The approach proposed here is totally new, efficient and promising to be incorporated in real-time Generalized Data Association - Multi Target Tracking systems (GDA-MTT) and provides an important result on the behavior of PCR5 with respect to Dempster's rule. The MatLab source code is provided in \cite{Dezert_F2K6}.
\end{abstract}


\section{Introduction}
The main purpose of information fusion is to produce reasonably aggregated, refined and/or complete granule of data obtained from a single or multiple sources with consequent reasoning process, consisting in using evidence to choose the best hypothesis, supported by it. Data Association (DA) with its main goal to partitioning observations into available tracks becomes a key function of any surveillance system. An issue to improve track maintenance performances of modern Multi Target Trackers (MTT) \cite{YBS_1990,Blackman_1999}, is to incorporate Generalized Data\footnote{Data being kinematics and attribute.} Association (GDA) in tracking algorithms \cite{Tchamova_2004}. At each time step, GDA consists in associating current (attribute and kinematics) measurements with predicted measurements (attributes and kinematics) for each target. GDA can be actually decomposed into two parts \cite{Tchamova_2004}: Attribute-based Data Association (ADA) and Kinematics-based Data Association (KDA). Once ADA is obtained, the estimation of the attribute/type of each target must be updated using a proper and an efficient fusion rule. This process is called {\it{attribute tracking}} and consists in combining information collected over time from one (or more) sensor to refine the knowledge about the possible changes of the attributes of the targets. We consider here the possibility that the attributes tracked by the system can change over time, like the color of a chameleon moving in a variable environment. In some military applications, target attribute can change since for example it can be declared as neutral at a given scan and can become a foe several scans later; or like in the example considered in this paper, a tracker can become mistaken when tracking several closely-spaced targets and thus could eventually {\it{track}} sequentially different targets observing that way a true sequence of different types of targets. In such case, although the attribute of each target is invariant over time, at the attribute-tracking level the type of the target committed to the (hidden unresolved) track varies with time and must be tracked efficiently to help to discriminate how many different targets  are hidden in the same unresolved track. Our motivation for attribute fusion is inspired from the necessity to ascertain the targets' types, information, that in consequence has an important implication to enhance the tracking performance. Combination rules are special types of the aggregation methods. To be useful, one system has to provide a way to capture, analyze and utilize through the fusion process the new available data (evidence) in order to update the current state of knowledge about the problem under consideration.\\

Dempster-Shafer Theory (DST) \cite{Shafer_1976} is one of the widely framework used in the area of target tracking when one wants to deal with uncertain information and take into account attribute data and/or human-based information into modern tracking systems. DST, thanks to belief functions, is well suited for representing uncertainty and combining information, especially in case of low conflicts between the sources (bodies of evidence) with high beliefs.
When the conflict increases\footnote{Which often occurs in Target Type Tracking problem as it will be showed in the sequel.} and becomes very high (close to 1), Dempster's rule yields unfortunately unexpected or what authors feel counter-intuitive results \cite{Zadeh_1979,DSmTBook_2004a}. Dempster's rule also presents difficulties in its implementation/programming because of unavoidable numerical rounding errors due to the finite precision arithmetic of our computers.\\

To overcome the drawbacks of Dempster's fusion rule and in the meantime extend the domain of application of the belief functions, we have proposed recently a new mathematical framework, called Dezert-Smarandache Theory (DSmT) with a new set of combination rules, among them the Proportional Conflict Redistribution no. 5 which proposes a sophisticated and efficient solution for information fusion as it will be showed further. The basic idea of DSmT is to work on Dedekind's lattice (called Hyper-Power Set) rather than on the classical power set of the frame as proposed in DST and, when needed, DSmT can also take into account the integrity constraints on elements of the frame, constraints which can also sometimes change over time with new knowledge. Hence DSmT deals with uncertain, imprecise and high conflicting information for static and dynamic fusion as well \cite{DSmTBook_2004a,Dezert _Smarandache_2005,Dezert _Smarandache_2006}.\\

In the next section we present briefly the basics of DST and DSmT. We recall the principles of Dempster's and PCR5 fusion rules. In section 3, we present the Target Type Tracking problem and examine two solutions to solve it; first solution being based on Dempster's rule and the second one based on PCR5.
In section 4, we evaluate both solutions on a very simple academic but checkable\footnote{Our MatLab source code is provided in \cite{Dezert_F2K6} to help the reader to check by him/herself the validity of our results.} example and provide a comparative analysis on Target Type Tracking performances obtained by Dempster's rule and PCR5. Concluding remarks are given in section 5.

\section{Basics on DST and DSmT}

Shafer's model, denoted here $\mathcal{M}^0(\Theta)$, in DST \cite{Shafer_1976} considers $\Theta=\{\theta_{1},\ldots,\theta_{n}\}$ as a finite set of $n$ exhaustive and exclusive elements representing the possible states of the world, i.e. solutions of the problem under consideration. $\Theta$ is called the {\it{frame of discernment}} by Shafer. In DSmT framework \cite{DSmTBook_2004a}, one starts with the free DSm  model $\mathcal{M}^f(\Theta)$ where $\Theta=\{\theta_{1},\ldots,\theta_{n}\}$ (called simply frame) is only assumed to be a finite set of $n$ exhaustive elements\footnote{The exclusivity assumption is not fundamental in DSmT because one wants to deal with elements which cannot be refined into precise finer exclusive elements - see \cite{DSmTBook_2004a} for discussion.}. If one includes some integrity constraints in $\mathcal{M}^f(\Theta)$, say by considering $\theta_1$ and $\theta_2$ truly exclusive (i.e. $\theta_1\cap\theta_2=\emptyset$), then the model is said {\it{hybrid}}. When we include all exclusivity constraints on elements of $\Theta$, $\mathcal{M}^f(\Theta)$ reduces to Shafer's model $\mathcal{M}^0(\Theta)$ which can be viewed actually as a particular case of DSm hybrid model. Between the free-DSm model and the Shafer's model, there exists a wide class of fusion problems represented in term of DSm hybrid models where $\Theta$ involves both fuzzy continuous hypothesis and discrete hypothesis.\\

Based on $\Theta$ and Shafer's model, the {\it{power set}} of $\Theta$, denoted $2^\Theta$, is defined as follows:
\begin{enumerate}
\item[1)] $\emptyset, \theta_1,\ldots, \theta_n \in 2^\Theta$.
\item[2)]  If $X, Y \in 2^\Theta$, then $X\cup Y$ belong to $2^\Theta$.
\item[3)] No other elements belong to $2^\Theta$, except those obtained by using rules 1) or 2).
\end{enumerate}

In DSmT and without additional assumption on $\Theta$ but  the exhaustivity of its elements (which is not a crucial assumption), we define the {\it{hyper-power set}}, i.e. Dedekind's lattice, $D^\Theta$ as follows:
\begin{enumerate}
\item[1')] $\emptyset, \theta_1,\ldots, \theta_n \in D^\Theta$.
\item[2')]  If $X, Y \in D^\Theta$, then $X\cap Y$ and $X\cup Y$ belong to $D^\Theta$.
\item[3')] No other elements belong to $D^\Theta$, except those obtained by using rules 1') or 2').
\end{enumerate}

When Shafer's model $\mathcal{M}^0(\Theta)$ holds, $D^\Theta$ reduces to the classical power set $2^\Theta$. Without loss of generality, we denotes $G^\Theta$ the general set on which will be defined the basic belief assignments (or masses), i.e. $G^\Theta=2^\Theta$ if Shafer's model is adopted whereas $G^\Theta=D^\Theta$ if some other (free or hybrid) DSm models are preferred depending on the nature of the problem. \\

From a frame $\Theta$, we define a (general) basic belief assignment (bba) as a mapping $m(.): 
G^\Theta \rightarrow [0,1]$ associated to a given source, say $s$, of evidence as 
\begin{equation}
m_s(\emptyset)=0 \qquad \text{and}\qquad \sum_{X\in G^\Theta} m_s(X) = 1 
\end{equation}
\noindent $m_s(X)$ is the gbba of $X$ committed by the source $s$. The elements of $G$ having a strictly positive mass are called {\it{focal elements}} of source $s$. The set $\mathcal{F}$ of all focal elements is the {\it{core}} (or {\it{kernel}}) of the belief function of the source $s$.\\

The belief and plausibility of any proposition $X\in G^\Theta$ are defined\footnote{The index of the source has been omitted for simplicity.} as:
\begin{equation}
\text{Bel}(X) \triangleq \sum_{\substack{Y\subseteq X\\ Y\in G^\Theta}} m(Y)
\ \text{and}\ \text{Pl}(X) \triangleq \sum_{\substack{Y\cap X\neq\emptyset \\ Y\in G^\Theta}} m(Y)
\end{equation}

These definitions remain compatible with the classical $\text{Bel}(.)$ and $\text{Pl}(.)$ functions proposed by Shafer in \cite{Shafer_1976} whenever Shafer's model is adopted for the problem under consideration since $G^\Theta$ reduces to $2^\Theta$.\\

A wide variety of rules exists for combining basic belief assignments  \cite{Sentz_2002,Smarandache_2005b,Smets_2005} and the purpose of this paper is not to browse and compare all these rules, but only show that the two main rules used with DST and DSmT approaches so far perform very differently on a very simple Target Type Tracking example. Let's now present the major differences between the two theories for combining sources of evidences. In DST framework, the fusion rule proposed by Glenn Shafer for combining several independent\footnote{While independence is a difficult concept to define in all theories managing epistemic uncertainty, we consider that two sources of evidence are independent (i.e. distinct and non-interacting) if each leaves one totally ignorant about the particular value the other will take.} source of evidences is Demspter's rule, while in DSmT, several rule have been proposed; mainly the DSm Hybrid rule, denoted (DSmH) which is a direct extension of Dubois \& Prade's rule of combination \cite{Dubois_1988} for working on $D^\Theta$ with dynamic fusion, and the recent and attractive Proportional Conflict Redistribution rule no. 5 (PCR5) \cite{Smarandache_2005c}. The DSm Hybrid rule consists just in transferring the partial conflicts onto the partial ignorances\footnote{Partial ignorance being the disjunction of elements involved in the partial conflicts.}, while as it will be seen, PCR5 redistributes the partial conflicting mass only to the elements
involved in that partial conflict and proportionally with respect to the masses each element put in the
partial conflict considering the conjunctive normal form of the partial conflict. No matter if the conflicting mass is big or small, PCR5  mathematically does a better redistribution of the conflicting mass than Dempster's rule and other rules since PCR5 goes backwards on the tracks of the conjunctive rule. For this reason, we only consider the PCR5 fusion rule in our comparative analysis with Dempster's rule. Both rules (Dempster's and PCR5) are mainly based on the conjunctive consensus operator defined for two-source case (which can be directly generalized for $N>2$ sources) by:
\begin{equation}
m_{12}(X) = 
\sum_{\substack{X_1,X_2\in G^\Theta\\ X_1\cap X_2=X}}m_{1}(X_1)m_{2}(X_2)
\label{Conj}
\end{equation}
\noindent
The {\it{degree of conflict}} (total conflict) between two sources represented by $m_1(.)$ and $m_2(.)$ is defined by 
 \begin{equation}
 k_{12}=\displaystyle\sum_{\substack{X_1,X_2\in  G^\Theta\\ X_1\cap X_2=\emptyset}} m_{1}(X_1) m_{2}(X_2)
\label{TotConf}
 \end{equation}
 \noindent
The total conflict $k_{12}$ is thus nothing but the sum of all partial conflicts. If $k_{12}$ is close to $0$, the bbas $m_1(.)$ and $m_2(.)$ are almost not in conflict, while if $k_{12}$ becomes close to $1$, the two sources are almost in total conflict.\\

From now on, we assume (without loss of generality) in the following presentation that the sources of evidence are equally reliable, otherwise a discounting preprocessing has to be applied first to each source according classical discounting method proposed in \cite{Shafer_1976}.

\subsection{Dempster's combination rule}

Dempster's rule has been proposed by Shafer in his Mathematical Theory of Evidence, usually referred also as Dempster-Shafer Theory \cite{Shafer_1976} to combine sources of evidence. Because the Shafer's model is used in DST, $G^\Theta=2^\Theta$. Dempster's combination rule for two sources is defined by $m_{D}(\emptyset)\triangleq 0$ and $\forall X \in 2^\Theta\setminus\{\emptyset\}$ by
\begin{equation}
m_{D}(X) = \frac{1}{1-k_{12}} m_{12}(X)
\label{eq:DSR}
 \end{equation}
\noindent
where $m_{12}(X)$ and $k_{12}$ are respectively defined by \eqref{Conj} and \eqref{TotConf}. Dempster's rule can be directly extended for the combination of $N$ independent and equally reliable sources of evidence and its major interest comes essentially from its commutativity and associativity properties \cite{Shafer_1976}. Dempster's rule corresponds to the normalized conjunctive rule by uniformly reassigning the mass of total conflict onto all focal elements through the conjunctive operator.
From theoretical point of view, Dempster's rule cannot be used only when $k_{12}=1$ because only in that case the division by zero occurs (which is mathematically not defined). From a practical point of view however, Dempster's rule is also difficult to use as soon as the conflict becomes very high (very close to one as in our applications) because the division by a very small number with finite precision processors yields rounding errors which can provide very instable/unexpected results. 

\subsection{PCR5 combination rule}

Instead of distributing equally the total conflicting mass onto elements of $2^\Theta$ as within Dempster's rule through the normalization step, or transferring the partial conflicts onto partial uncertainties as within DSmH rule, the idea behind the Proportional Conflict Redistribution rules \cite{Smarandache_2005c} is to transfer conflicting masses (total or partial) proportionally to non-empty sets involved in the model according to all integrity constraints. The general principle of PCR rules is then to :
\begin{enumerate}
\item calculate the conjunctive rule of the belief masses of sources ;
\item calculate the total or partial conflicting masses ;
\item redistribute the conflicting mass (total or partial) proportionally on non-empty sets involved in the model according to all integrity constraints. 
\end{enumerate}
The way the conflicting mass is redistributed yields actually to several versions of PCR rules \cite{Smarandache_2005c}. These PCR fusion rules work for any degree of conflict in $[0, 1]$, for any DSm models (Shafer's model, free DSm model or any hybrid DSm model) and both in DST and DSmT frameworks for static or dynamical fusion problems. We just now present only the most sophisticated proportional conflict redistribution rule no. 5 (PCR5) since this rule is what we feel the most efficient PCR fusion rule proposed\footnote{A new PCR6 rule has been developed very recently by Martin and Osswald \cite{Martin_2006} but will not be presented and discussed here since it coincides with PCR5 for the two-source case in our application.} so far.\\

The PCR5 combination rule for only two sources\footnote{A general expression of PCR5 for an arbitrary number ($s>2$) of sources can be found in \cite{Dezert _Smarandache_2006}.} is defined by \cite{Smarandache_2005c}: $m_{PCR5}(\emptyset)=0$ and $\forall X\in G^\Theta\setminus\{\emptyset\}$
\begin{equation}
m_{PCR5}(X)=m_{12}(X) +\sum_{\substack{Y\in G^\Theta\setminus\{X\} \\ c(X\cap Y)=\emptyset}} 
[\frac{m_1(X)^2m_2(Y)}{m_1(X)+m_2(Y)} +
 \frac{m_2(X)^2 m_1(Y)}{m_2(X)+m_1(Y)}]
   \label{eq:PCR5}
 \end{equation}
\noindent
where $m_{12}(X)$ corresponds to the conjunctive consensus on $X$ between the two sources and where all denominators are different from zero and  $c(X)$ is the canonical form\footnote{The canonical form is introduced here explicitly in order to improve the original formula given in \cite{DSmTBook_2004a} for preserving the neutral impact of the vacuous belief mass $m(\Theta)=1$ within complex hybrid models. Actually all propositions involved in formulas are expressed in their canonical form, i.e. conjunctive normal form, also known as conjunction of disjunctions in Boolean algebra, which is unique.} of $X$, i.e. its simplest form (for example if $X=(A\cap B)\cap (A\cup B\cup C)$, $c(X)=A\cap B$). If a denominator is zero, that fraction is discarded.\\

No matter how big or small is the conflicting mass, PCR5 mathematically does a better redistribution of
the conflicting mass than Dempster's rule and other rules since PCR5 goes backwards on the tracks of the conjunctive rule and redistributes the partial conflicting masses only to the sets involved in the conflict and proportionally to their masses put in the conflict, considering the conjunctive normal form of the partial conflict. PCR5 is quasi-associative and preserves the neutral impact of the vacuous belief assignment. \\

In short summary, the main differences between DST and DSmT are (1) the model on which one works with, and (2) the choice of the combination rule.

\section{The Target Type Tracking Problem }

\subsection{Formulation of the problem}

The Target Type Tracking Problem can be simply stated as follows:
\begin{itemize}
\item Let $k=1,2,...,k_{max}$ be the time index and consider $M$ possible target types $T_i\in \Theta=\{\theta_1,\ldots,\theta_M\}$ in the environment; for example $\Theta=\{Fighter,Cargo\}$ and $T_1\triangleq Fighter$, $T_2\triangleq Cargo$; or $\Theta=\{Friend,Foe,Neutral\}$, etc.
\item at each instant $k$, a target of true type $T(k)\in \Theta$ (not necessarily the same target) is observed by an attribute-sensor (we assume a perfect target detection probability here).
\item the attribute measurement of the sensor (say noisy Radar Cross Section for example) is then processed through a classifier which provides a decision $T_d(k)$ on the type of the observed target at each instant $k$.
\item The sensor is in general not totally reliable and is characterized by a $M\times M$ confusion matrix 
$$\mathbf{C}=[c_{ij}=P(T_d=T_j|TrueTargetType=T_i)]$$
\end{itemize}

Question: How to estimate $T(k)$ from the sequence of declarations done by the unreliable classifier up to time $k$, i.e. how to build an estimator $\hat{T}(k)=f(T_d(1),T_d(2),\ldots,T_d(k))$ of $T(k)$ ?

\subsection{Proposed issues}

We propose in this work two methods for solving the Target Type Tracking Problem. Both methods assume same Shafer's model for the frame of Target Types $\Theta$ and also use the same information (vacuous belief assignment as prior belief and same sequence of {\it{measurements}}, i.e. same set of classifier declarations to get a fair comparative analysis). The proposed issues are based on the combination of belief functions. \\

The principle of our estimators is based on the sequential combination of the current basic belief assignment (drawn from classifier decision, i.e. our {\it{measurements}}) with the prior bba estimated up to current time from all past classifier declarations. In the first approach, the Demspter's rule is used for estimating the current Target type, while in the second approach we use PCR5.\\

\noindent
Here is how our Target Type Tracker (TTT) works:

\begin{itemize}
\item a) Initialization step (i.e. $k=0$). Select the target type frame $\Theta=\{\theta_1,\ldots,\theta_M\}$ and set the prior bba $m^{-}(.)$ as vacuous belief assignment, i.e $m^{-}(\theta_1\cup\ldots\cup\theta_M)=1$ since one has no information about the first target type that will be observed.
\item b) Generation of the current bba $m_{obs}(.)$ from the current classifier declaration $T_d(k)$ based on attribute measurement. At this step, one takes $m_{obs}(T_d(k))=c_{T_d(k)T_d(k)}$ and all the unassigned mass $1-m_{obs}(T_d(k))$ is then committed to total ignorance $\theta_1\cup\ldots\cup\theta_M$.
\item c) Combination of current bba $m_{obs}(.)$ with prior bba $m^{-}(.)$ to get the estimation of the current bba $m(.)$. Symbolically we will write the generic fusion operator as $\oplus$, so that $m(.)=[m_{obs}\oplus m^{-}](.)=[m^{-}\oplus m_{obs}](.)$. The combination $\oplus$ is done according either Demspter's rule (i.e. $m(.)=m_D(.)$) or PCR5 rule (i.e. $m(.)=m_{PCR5}(.)$).
\item d) Estimation of True Target Type is obtained from $m(.)$ by taking the singleton of $\Theta$, i.e. a Target Type, having the maximum of belief (or eventually the maximum Pignistic Probability\footnote{We don't provide here the results based on Pignistic Probabilities since in our simulations the conclusions are unchanged when working with max. of belief or max. of Pign. Proba.} \cite{DSmTBook_2004a}).
\item e) set $m^{-}(.)=m(.)$; do $k=k+1$ and go back to step b).
\end{itemize}

\section{Simulations results }

In order to evaluate the performances of both estimators and have a fair comparative analysis of the Dempster's and PCR5 fusion rules, we did a set of Monte-Carlo simulations on a very simple scenario for a 2D Target Type frame, i.e. $\Theta=\{(F)ighter,(C)argo\}$ for two classifiers, a good one $C_1$ and a poor one $C_2$ corresponding to the following confusion matrices:
$$\mathbf{C}_1=
\begin{bmatrix}
0.95 & 0.05\\
0.05 & 0.95
\end{bmatrix}
\quad
\text{and}\quad
\mathbf{C}_2=
\begin{bmatrix}
0.75 & 0.25\\
0.25 & 0.75
\end{bmatrix}
$$

In our scenario we consider that there are two closely-spaced targets: one cargo and one fighter. Due to circumstances, attribute measurements received are predominately from one or another, and both target generates actually one single (unresolved kinematics) track. In the real world, the tracking system should in this case maintain two separate tracks: one for cargo and one for fighter, and based on the classification, allocate the measurement to the proper track. But in difficult scenario like this one, there is no way in advance to know the true number of targets because they are unresolved and that's why only a single track is maintained. Of course, the single track can further be split into two separate tracks as soon as two different targets are declared based on the attribute tracking. This is not the purpose of our work however since we only want to examine how work PCR5 and Dempster's rules for Target Type Tracking. To simulate such scenario, a true Target Type sequence over 120 scans was generated according figure 1 below. The sequence starts with the observation of  a Cargo Type (i.e. we call it Type 2) and then the observation of the Target Type switches three times onto Fighter Type (we call it Type 1) during different time duration (20 s, 10 s and 5 s). As a simple analogy, tracking the target type changes committed to the same (hidden unresolved) track can be interpreted as tracking color changes of a chameleon moving in a tree on its leaves and on its trunk.\\

\begin{figure}[!hbtp]
\centering
\includegraphics[width=8cm]{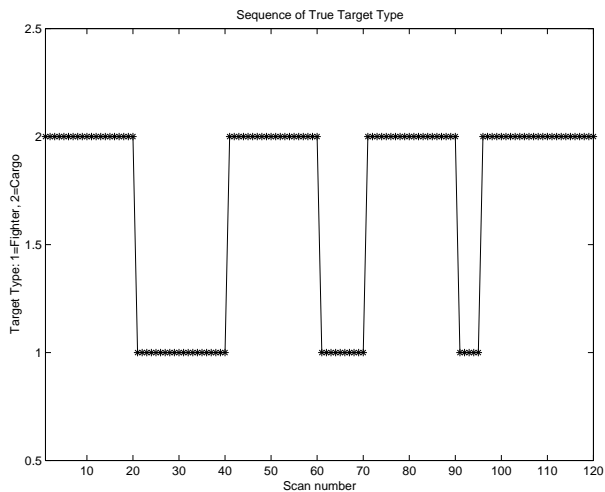}
\caption{Sequence of True Target Type (Groundthruth)}
 \label{fig:1}
 \end{figure}
 
 Our simulation consists in 1000 Monte-Carlo runs and we compute and show in the sequel the averaged performances of the two fusion rules. At each time step $k$ the decision $T_d(k)$ is randomly generated according to the corresponding row of the confusion matrix of the classifier given the true Target Type (known in simulations). Then the algorithm presented in the previous section is applied. The MatLab source code of our simulation is provided in \cite{Dezert_F2K6} for convenience.
 
\subsection{Results for classifier 1}
 
\begin{figure}[h]
\centering
\includegraphics[width=8cm]{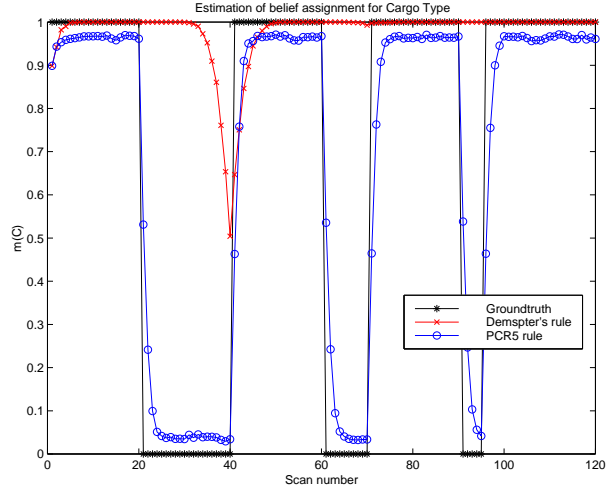}
\caption{Belief mass for Cargo Type for $C_1$}
 \label{fig:2}
 \end{figure}

\begin{figure}[h]
\centering
\includegraphics[width=8cm]{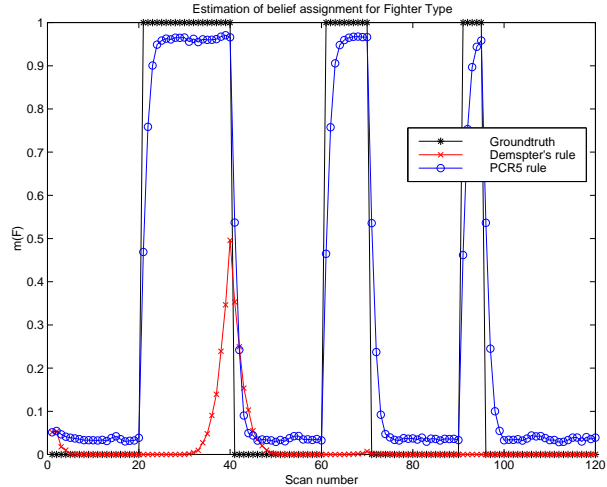}
\caption{Sequence of beliefs for Fighter Type for $C_1$}
 \label{fig:3}
 \end{figure}

Figures \ref{fig:2} and \ref{fig:3} show the belief masses obtained by our Target Type Trackers based on Demspter's rule (red curves -x-) and PCR5 rule (blue curves -o-). It can be seen that the TTT based on Dempster's rule and for a very good classifier is unable to track properly the quick changes of target type. This phenomenon is due to the too long integration time necessary to the Demspter's rule for recover true belief estimation. \\

Demspter's rule presents a very long latency delay (about 18 s as we can see during the first type switch) when almost all the basic belief mass is committed onto only one element of the frame. PCR5 rule can quickly detect the type changes and  properly re-estimates the belief masses contrariwise to Dempster's rule. So in this configuration the TTT based on Demspter's rule works almost blindly since it is unable to detect the fighter in most of scans where the true target type is a Fighter. Figures \ref{fig:2} and \ref{fig:3} show clearly the efficiency of PCR5 rule with respect to Demspter's rule.


\subsection{Results for classifier 2}

\begin{figure}[h]
\centering
\includegraphics[width=8cm]{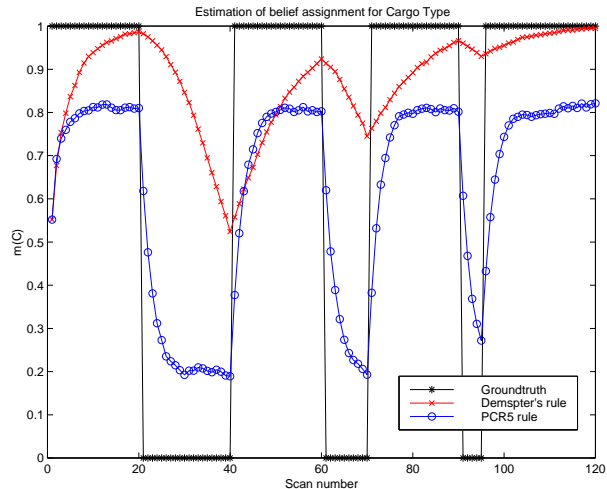}
\caption{Belief mass for Cargo Type for $C_2$}
 \label{fig:4}
 \end{figure}

\begin{figure}[h]
\centering
\includegraphics[width=8cm]{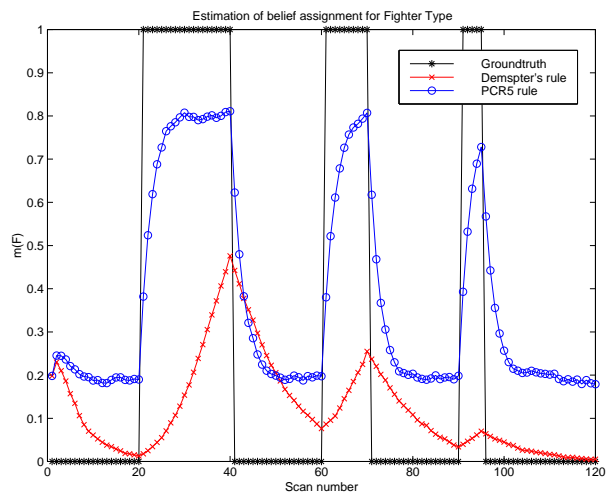}
\caption{Sequence of beliefs for Fighter Type for $C_2$}
 \label{fig:5}
 \end{figure}

Figures \ref{fig:4} and \ref{fig:5} show the belief masses obtained by our TTT based on Demspter's rule (red curves) and PCR5 rule (blue curves) with classifier 2. Paradoxically, we can observe that the Demspter's rule seems to work better with a poor classifier than with a good one, because we can see from the red curves that Dempster's rule in that case produces small change detection peaks (with always an important latency delay although). This phenomenon is actually not so surprising and comes from the fact that the belief mass of the true type has not well been estimated by Dempster's rule (since the mass is not so close to its extreme value) and thus the bad estimation of Target Type facilitates the ability of Dempster's rule to react to new incoming information and detect changes. When from Demspter's rule, one obtains an over-confidence onto only one focal element of the power-set, it then becomes very difficult for the Dempster's rule to readapt automatically, efficiently and quickly to any changes of the state of the nature which varies with the time and this behavior is very easy to check either analytically or through simple simulations. The major reason for this unsatisfactory behavior of Dempster's rule can be explained with its main weakness: counterintuitive averaging of strongly biased evidence, which in the case of poor classifier is not valid. We can see the ability of PCR5 to track Target Type and detect the short Type changes even when using a poor classifier. More examples with sensitivity analysis including results for other fusion rules can be found n \cite{Dezert_2006Book2}.

\section{Conclusions}

Two Target Type Trackers (TTT) have been proposed and compared in this paper. Our trackers are based on two combinational rules for temporal attribute data fusion for target type estimation: 1) Dempster's rule drawn from Dempster-Shafer Theory (DST) and 2) PCR5 rule drawn from Dezert-Smarandache Theory (DSmT). Our comparative analysis shows through a very simple scenario and Monte-Carlo simulation that PCR5 allows a very efficient Target Type Tracking, reducing drastically the latency delay for correct Target Type decision, while Dempster's rule demonstrates risky behavior, keeping indifference to the detected target type changes. The temporal fusion process utilizes the new knowledge in an incremental manner and hides the possibility for arising bigger conflicts between the new coming and the previous updated evidence. Dempster's rule cannot detect quickly and efficiently target type changes, and thus to track them correctly. It hides the risk to produce counter-intuitive and non adequate results. Our PCR5-based Target Type Tracker is totally new, efficient and promising to be incorporated in real-time Generalized Data Association - Multi Target Tracking systems (GDA-MTT). It provides an important result on the behavior of PCR5 with respect to Dempster's rule.

\end{document}